\documentclass[]{llncs}
\usepackage[width=122mm,left=12mm,paperwidth=146mm,height=193mm,top=12mm,paperheight=217mm]{geometry}
\begin{document}

\title{Observation Centric and Central Distance Recovery on Sports Player Tracking} 

\author{Hsiang-Wei Huang, Cheng-Yen Yang, Jenq-Neng Hwang \\Pyong-Kun Kim, Kwangju Kim, Kyoungoh Lee}
\institute{University of Washington, Seattle \\
Electronics and Telecommunications Research Institute, Korea}
\maketitle

\begin{abstract}
Multi-Object Tracking over humans has improved rapidly with the development of object detection and re-identification. However, multi-actor tracking over humans with similar appearance and non-linear movement can still be very challenging even for the state-of-the-art tracking algorithm. Current motion-based tracking algorithms often use Kalman Filter to predict the motion of an object, however, its linear movement assumption can cause failure in tracking when the target is not moving linearly. And for multi-players tracking over the sports field, because the players in the same team are usually wearing the same color of jersey, making re-identification even harder both in the short term and long term in the tracking process. In this work, we proposed a motion-based tracking algorithm and three post-processing pipelines for three sports including basketball, football, and volleyball, we successfully handle the tracking of the non-linear movement of players on the sports fields. Experiments result on the testing set of ECCV DeeperAction Challenge SportsMOT Dataset demonstrate the effectiveness of our method, which achieves a HOTA of 73.968, ranking 3rd place on the 2022 Sportsmot workshop final leaderboard.
\end{abstract}

\section{Introduction}
Multi-object tracking is a fundamental task in computer vision, aiming to associate objects bounding boxes and keep track of all the identities in video sequences.
Most of the multi-object tracking datasets mainly focus on pedestrians in crowded street scenes (e.g., MOT17/20) [1]. In these scenes, most of the pedestrians' movement is slow and linear and thus easy to predict, besides that, the appearance of each identity is easy to distinguish, making the re-identification in the tracking process easier. However, there is a lack of multi-object tracking algorithms that can successfully handle the non-linear movement of sports players and the challenge of similar appearances between players on the sports field scenes. For this purpose, we introduced an observation-centric and central distance recovery tracking algorithm that can handle the non-linear movement of players on the sports field, and also use appearance ReID post-processing to deal with the fragment tracklets during tracking.

\section{Related Works}

\subsection{Location and Motion Based Object Tracking}
Modern object tracking algorithms usually follow the paradigm of tracking by detection. Several motion-based tracking algorithms adopt the Kalman filter [2] to formulate the moving trajectories of the target with the detections from the object detector. However, the linear assumptions of the Kalman filter can fall short when the target is not moving linearly. Observation-Centric SORT [3] includes moving direction similarity between detection and tracklets into the Hungarian Algorithm [4] cost and reforming the trajectories of the re-identify target to prevent error accumulations in the Kalman filter update process. These methods show effectiveness when dealing with the non-linear movements of objects, achieving state-of-the-art tracking performance on several multi-object tracking datasets.

\subsection{Appearance Based Object Tracking}
With the fast development in the appearance feature extractor, some tracking algorithms incorporate target appearance during association [5,6] and utilizes the appearance as a clue for identity recognition. However, most of these appearance cues-based algorithms often fall short in many cases especially when targets are occluded, when the scene is very crowded or the targets are sharing similar appearances.

\section{Proposed Method}
Several challenges need to be tackled in the tracking of sports players. First, is the non-linear movement of the players on the court. Given the high intensity of sports like volleyball, basketball, and football, players need to sprint, jump and change directions in a short time during the game, causing the movement to be unpredictable. Second, is the heavy occlusion problem during the sports game. The players will cluster together during the sports game when some particular situations happened like grabbing the rebound in basketball or blocking in volleyball. When this happened, it causes detection to be unreliable due to occlusion, and thus causes the tracking performance to drop if we do not take care of the recovery of the tracklets with lost detection carefully. Third, in the sports video clips, a player can go out and re-enter the camera view again after several seconds, when this happened, we need to re-identify the player with the correct identity. However, given the similar appearance between those team players within the same team, it is difficult to re-identify them correctly by using their appearance. In our work, we proposed several methods to deal
with these three challenges.

\subsection{Observation-Centric Tracking}
The non-linear movement is causing big trouble during the tracking process. To tackle this problem, we use Observation-Centric SORT (OCSORT) as our main tracking algorithm. Its Observation-centric Online Smoothing strategy helps deal with the non-linear movement of the target by rebuilding a virtual trajectory when a lost target is associated again after a period of being untracked, thus preventing error accumulation in the Kalman filter. Because the players on the sports field can change moving direction in a short time, the Observation-Centric Momentum in OCSORT also helps reduce the tracking error caused by sudden direction changes. Lastly, after the association stage is finished, OCSORT performs Observation-Centric Recovery, trying to associate the last observation of an unassociated track to the detections on the new-coming time step. This strategy helps to reduce the generation of a new tracklet, which is usually abnormal given the fixed number of players on the sports field.

\subsection{Central Distance Recovery}
After the association stage is done in OCSORT, there is still a chance that observation-centric recovery failed to successfully recover the identity of the targets and initiate a new tracklet. This is mainly due to observation-centric recovery being based on using bounding box IoU as the Hungarian assignment cost, while the unassociated tracklets and the unmatched detections sometimes do not necessarily share an overlapping region given the fast-moving speed of sports players. To deal with this, we use the Euclidean distance between detections and the last observation of unassociated tracklets to conduct observation-centric recovery again, as this stage is based on bounding boxes' central distance, we call this process “central distance recovery”.

\subsection{Tracklets Association Post-Processing}
After the tracking stage is finished, the result usually ended up in a bigger number of identities than the number of identities in the ground truth. This is because when a player re-enters the camera view, the tracking algorithm usually treats the player as a new identity and does not re-identify the player. To deal with the player re-entry problem, we incorporate different strategies in three sports scenes according to the total number of sports players, the size of the sports field, and several other sports characteristics.
\subsubsection{Post-Processing in Basketball}
\ 
\newline The total number of players in a basketball game is 10, we use this as the main constraint to conduct the post-processing. Due to a large amount of player's camera re-entry in basketball compared to other sports, it is necessary to incorporate person re-identification in the post-processing stage. After we get the preliminary tracking result, we initiate the first 10 tracklets we get as the ten main players on the court. We keep updating the tracklets appearance feature using exponential moving averages. Whenever a player leaves the camera view, we put the player's tracklet into a candidate queue. And whenever a new tracklet (player) enters the camera view, we associate the player to one of the tracklets in the candidate queue that shares the highest cosine similarity between both. 
\subsubsection{Post-Processing in Football}
\ 
\newline The search space of candidates in football is much bigger compared to other sports because of the bigger number of players in a football game and the lower ratio of players inside the camera view to the total number of players. Thus it is necessary to incorporate the position as a constraint during the ReID process. In football tracking, due to the uncertainty in the number of identities in a video clip, we decided not to use the number of total players as a constraint, instead, we only try to associate fragment tracklets together based on their appearance similarity and entry/exit position. There is a total of three rounds of association based on appearance similarity and location between tracklets. The prerequisites for two tracklets to be associated together are first based on their disappear and reappear location. The threshold for the location distance for two tracklets to be associated is determined by the number of frames between a tracklet disappearing and reappearing in the camera view, which means that if a tracklet disappears and reappears in the camera view in a short amount of time, the distance threshold for their location will be small given the moving distance should not be far in such a short time, and vice versa for a longer disappearing time. After passing the threshold of location distance, we then try to compare the cosine similarity based on the two tracklets' appearance. We calculate two tracklets' average frame-based embedding features distance as their final distance. If the final distance is smaller than the embedding threshold, we consider these two tracklets the same identity. We use three rounds of association based on a different matching threshold of appearance similarity.
\subsubsection{Post-Processing in Volleyball}
\
\newline Compared to basketball and football, volleyball player usually stays in the camera view throughout the entire video clip, and even they disappear, they reappear in a very close location. Due to the above reason, we do not incorporate appearance in the post-processing of volleyball. We use a similar strategy to basketball post-processing, we limited the number of tracklets to 12 players, and try to re-associate disappearing players to candidates only based on the distance of their disappear and reappear location.
\subsubsection{Interpolation}
\ 
\newline After the ReID post-processing part is done, we use linear interpolation as our last step to produce the final tracking results. 

\section{Experiments and Results}
\subsection{Datasets}
We use the training sets from SportsMOT for detector and ReID model training. The training set contains 45 video clips from 3 categories (i.e., basketball, football, and volleyball), which are collected from Olympic Games, NCAA Championship, and NBA games on YouTube. Only the search results with 720P resolution, 25 FPS, and official recordings are downloaded. All of the selected videos are cut into clips of average 485 frames manually, in which there is no shot change.
\subsection{Implementation Details}
\subsubsection{Detector}
\ 
\newline We use YOLOX [7] as our detector due to its high accuracy and fast inference speed. For the pretrained weight, we use the COCO pretrained YOLOX-X model provided by the official GitHub repositories of YOLOX. We train the model with Sportsmot training set for 80 epochs, following the YOLOX-X default training process of ByteTrack's [8] official GitHub repositories. The training duration takes around 8 hours on 4 tesla V100 GPUs.
\subsubsection{Observation-Centric SORT}
\ 
\newline We keep the original configuration of OCSORT, using 0.1 detection confidence threshold, 0.3 IoU threshold, 0.7 track threshold, and a maximum tracklet age of 30 frames for all of the sports.

\subsubsection{Central Distance Recovery}
\ 
\newline The central distance recovery threshold is different based on the sports type. We set basketball's distance threshold as 200, football's distance threshold as 80, and volleyball's distance threshold as 80. The choice of threshold is based on the evaluation performance on the Sportsmot testing set.
\subsubsection{Person ReID}
\ 
\newline We are using OSNet[9] as our backbone network for feature extraction. The model is trained with Sportsmot training set for 10 epochs, using Adam optimizer with 0.0003 learning rate.
\subsubsection{Basketball Post-Processing Setting}
\ 
\newline In the post-processing of basketball, we limited the number of tracklets to 10, just like the number of players on the court, unless more than 10 detections appear at the same time, we do not initiate new tracklets. The re-identification of players is based on the cosine similarity of their embedding features.
\subsubsection{Football Post-Processing Setting}
\ 
\newline In football, considering the ratio of players inside and outside the camera, we use the cosine similarity and also players' position to conduct re-identification for camera re-entry players. There are three rounds of the association stage. The association between two tracklets needs to pass through a threshold of tracklet position distance before they have a chance to be associated. The distance threshold is based on their disappear and reappear time gap, for those tracklets that have a time gap of fewer than 100 frames, the distance threshold is 100. For the tracklets that share a time gap between 100 to 500 frames, the distance threshold is 250, and for those tracklets that share a time gap bigger than 500 frames, the distance threshold is set to 400. In three rounds of association, we try to associate as many as tracklets we can in a greedy style. For the first round of association, if the cosine distance of two tracklets is smaller than 0.1, we treat them as the same identity and conduct association. For the second round, the cosine distance threshold is 0.2, and for the third round is 0.4.
\subsubsection{Volleyball Post-Processing Setting}
\ 
\newline Due to the relatively low number of player camera re-entry cases in volleyball compared to basketball and football, the search space for re-entry players is small. So the post-processing of volleyball is simply based on their disappear and reappear position. For the re-entry player, we select the player in the re-associate candidate who has the closest distance between the disappearing position of candidates and reappears position of the new player for re-identification if their distance is lower than a threshold of 400.

\subsection{Evaluation Results}
\
\newline The 2022 ECCV DeeperAction Challenge - SportsMOT Track on Multi-actor Tracking competition is ranked according to the HOTA [10] performance. In contrast to MOTA [11], HOTA maintains a balance between the accuracy of object detection and association. The original OCSORT has a performance of 67.107 in HOTA, after the incorporation of central-distance recovery, the HOTA improves to 71.764. After the ReID post-processing stage, we achieve 73.968 in HOTA, 63.460 in AssA, 86.316 in DetA, 94.832 in MOTA, 78.271 in IDF1, 2754 in IDS, and 3592 in Frag, showing the effectiveness of our method.

\section{Conclusions}
In this paper, we modify the motion-based observation-centric SORT with an extra central distance recovery stage, improving the performance without adding too much computational cost and also keeping the algorithm online, successfully tackle down the challenges of non-linear movement during tracking. We also propose a ReID post-processing stage for each sport according to the sports characteristics. Our final performance achieves 73.968 in HOTA, ranking 3rd place among all of the teams in the 2022 ECCV DeeperAction Challenge - SportsMOT Track on Multi-actor Tracking. 

\section{References}
\begin{enumerate}
\item A. Milan, L. Leal-Taixe, I. Reid, S. Roth, and K. Schindler. Mot16: A benchmark for multi-object tracking. arXiv preprint arXiv:1603.00831, 2016.
\item R. E. Kalman. A new approach to linear filtering and prediction problems. J. Fluids Eng., 82(1):35–45, 1960.
\item CAO, Jinkun, et al. Observation-Centric SORT: Rethinking SORT for Robust Multi-Object Tracking. arXiv preprint arXiv:2203.14360, 2022.
\item H. W. Kuhn. The hungarian method for the assignment problem. Naval research logistics quarterly, 2(1-2):83–97, 1955.
\item N. Wojke, A. Bewley, and D. Paulus. Simple online and realtime tracking with a deep association metric. In 2017 IEEE international conference on image processing (ICIP), pages 3645–3649. IEEE, 2017
\item Y. Zhang, C. Wang, X. Wang, W. Zeng, and W. Liu. Fairmot: On the fairness of detection and re-identification in multiple object tracking. arXiv preprint arXiv:2004.01888, 2020.
\item Z. Ge, S. Liu, F. Wang, Z. Li, and J. Sun. Yolox: Exceeding yolo series in 2021. arXiv preprint arXiv:2107.08430, 2021.
\item Yifu Zhang, Peize Sun, Yi Jiang, Dongdong Yu, Zehuan Yuan, Ping Luo, Wenyu Liu, and Xinggang Wang. Bytetrack: Multi-object tracking by associating every detection box. arXiv preprint arXiv:2110.06864, 2021.
\item Zhou, Kaiyang, et al. "Omni-scale feature learning for person re-identification." Proceedings of the IEEE/CVF International Conference on Computer Vision. 2019.
\item J. Luiten, A. Osep, P. Dendorfer, P. Torr, A. Geiger, L. LealTaixe, and B. Leibe. Hota: A higher order metric for evaluating multi-object tracking. International journal of computer vision, 129(2):548–578, 2021.
\item K. Bernardin and R. Stiefelhagen. Evaluating multiple object tracking performance: the clear mot metrics. EURASIP Journal on Image and Video Processing, 2008:1–10, 2008.
\end{enumerate}
\end{document}